\documentclass[10pt,twocolumn,letterpaper]{article}

\usepackage[pagenumbers]{wacv} 

\definecolor{wacvblue}{rgb}{0.21,0.49,0.74}
\usepackage[pagebackref,breaklinks,colorlinks,allcolors=wacvblue]{hyperref}

\def\wacvPaperID{312} 
\def\confName{WACV}
\def\confYear{2026}

\title{Sketch-guided Cage-based 3D Gaussian Splatting Deformation}

\author{
Tianhao Xie\textsuperscript{1}\quad
Noam Aigerman\textsuperscript{2,3} \quad
Eugene Belilovsky\textsuperscript{1,3} \quad
Tiberiu Popa\textsuperscript{1} \\
\textsuperscript{1}Concordia University, Montr\'eal, Canada \quad
\textsuperscript{2}Universit\'e de Montr\'eal, Montr\'eal, Canada \quad
\\
\textsuperscript{3}Mila, Montr\'eal, Canada \quad
}

\begin{document}
\twocolumn[{%
\renewcommand\twocolumn[1][]{#1}%
\maketitle
\includegraphics[width=\linewidth]{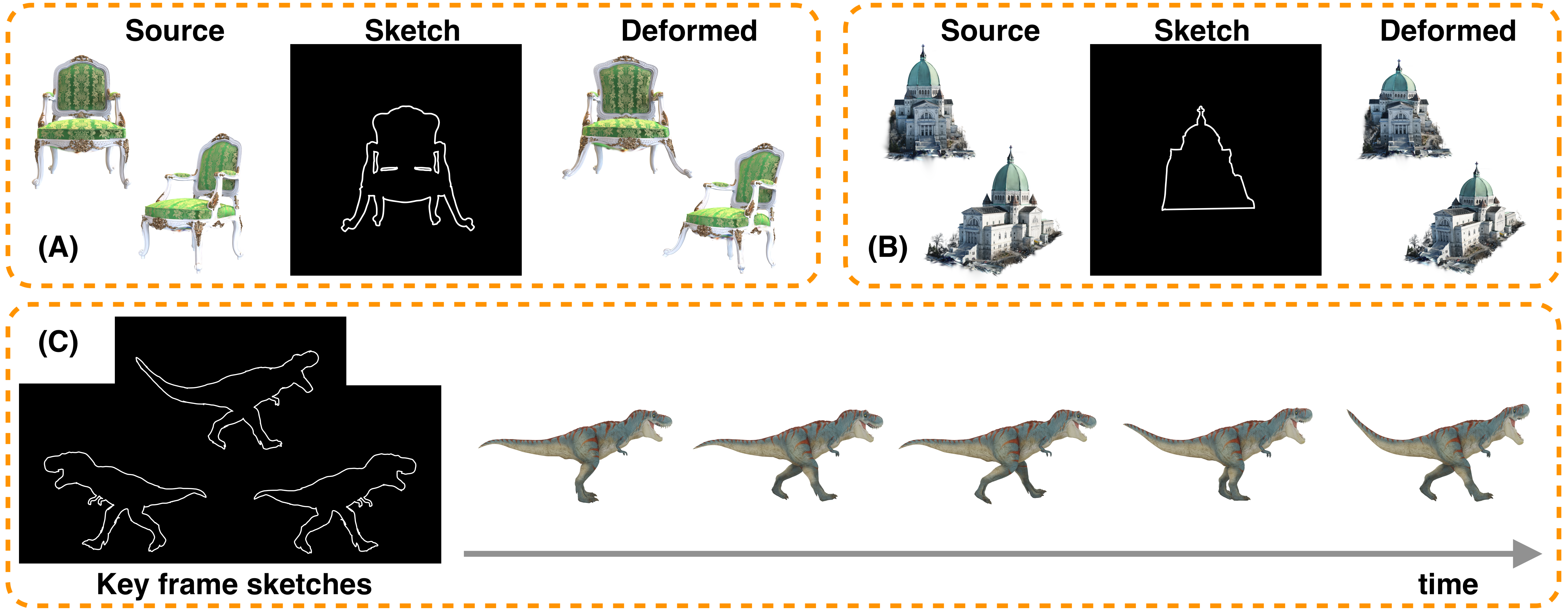}
\captionof{figure}{Deformations of  3D Gaussian splats (GS) using our sketch-guided deformation method. Given a 3D GS scene, the user can deform the 3D GS by drawing a deformed silhouette sketch of a single view. (A) and (B) show examples with synthetic data and real-world large-scale data, respectively. We can also produce an animation of a of 3D GS by few ($\geq 2$) keyframe sketches, as shown in (C).\vspace{1em}}
\label{fig:teaser}
}]
\begin{abstract}
3D Gaussian Splatting (GS) is one of the most promising novel 3D representations that has received great interest in computer graphics and computer vision. While various systems have introduced editing capabilities for 3D GS, such as those guided by text prompts, fine-grained control over deformation remains an open challenge. In this work, we present a novel sketch-guided 3D GS deformation system that allows users to intuitively modify the geometry of a 3D GS model by drawing a silhouette sketch from a single viewpoint. Our approach introduces a new deformation method that combines cage-based deformations with a variant of Neural Jacobian Fields, enabling fine-grained control.
Additionally, it leverages 2D diffusion priors and ControlNet to ensure the generated deformations are semantically plausible. Through a series of experiments, we demonstrate the effectiveness of our method and showcase its ability to animate static 3D GS models as one of its applications.
\end{abstract}    
\section{Introduction}
\label{sec:intro}
\begin{figure*}[t]
    \centering
    \includegraphics[width=\linewidth]{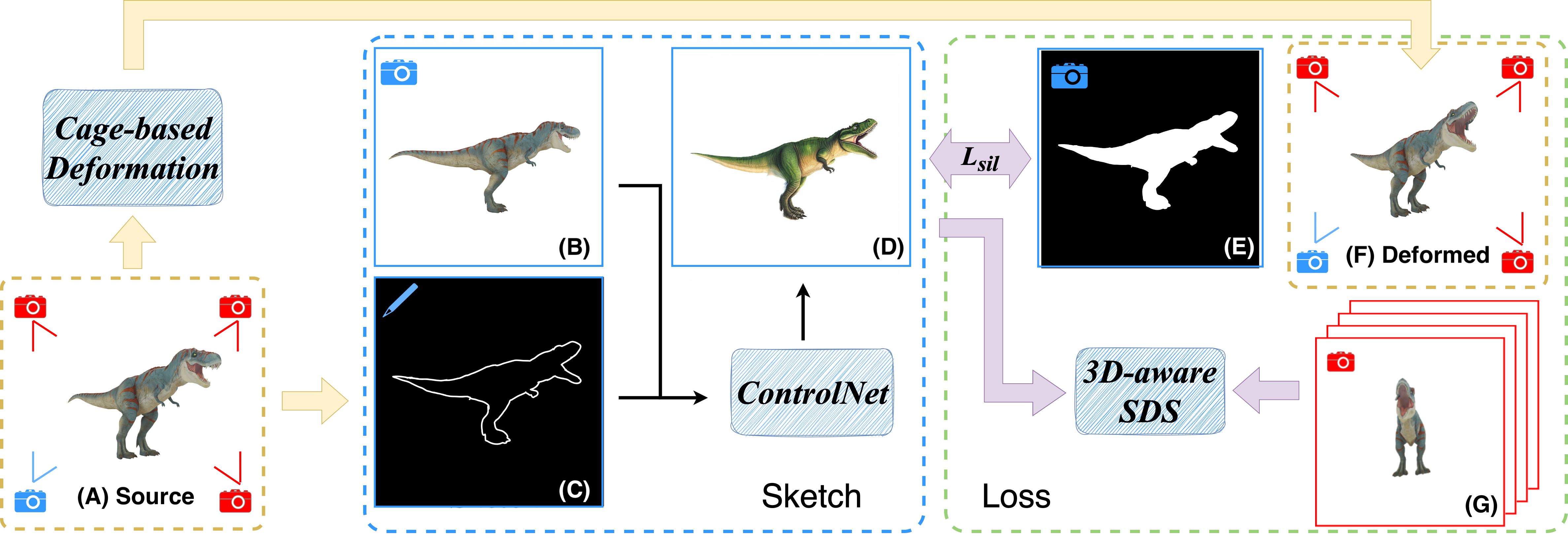}
    \caption{Overview of our method. We start with a 3D GS model (A). Users can select one specific sketch view (B) to draw an edited silhouette sketch (C). This user-drawn sketch (C) and rendering from the chosen view (B) are fed into ControlNet~\cite{zhang2023adding} resulting in a deformed reference image (D). 
    We then optimize the cage of the source 3D GS model (A) using two losses: (1) a silhouette loss $L_{sil}$ between the silhouette (E) of the deformed GS model (F) and the silhouette of the generated reference image (D), and (2) a 3D-aware SDS loss from $4$ random views (G) conditioned on the reference image (D).
    }
    \label{fig:method_overview}
\end{figure*}
Editing of 3D models and shapes often arises in computer graphics, computer animation, and geometric modeling. This editing is usually carried out by \textit{deforming} the 3D models. In this work, we aim to provide such editing-through-deformation capabilities for one of the most promising novel representations of 3D geometry - 3D Gaussian Splatting (GS)~\cite{kerbl20233d}, which offers great real-time novel view rendering ability and better photorealistic reconstruction than previous methods. 

Similarly to other geometric representations such as NeRFs~\cite{mildenhall2020nerf} and triangle meshes, various control mechanisms were proposed for editing GS, such as text prompts~\cite{chen2024gaussianeditor,wang2024gaussianeditor,chen2024dge,wu2024gaussctrl} and video priors~\cite{ren2023dreamgaussian4d,ling2024align}. Unfortunately, these types of controls are designed for broad, high-level edits (ones within the capabilities of a novice user), without enabling fine-grained control over the deformation. On the other hand, some investigation has been made into more direct, geometrical editing, e.g., via physics-based simulation~\cite{xie2024physgaussian} - this again offers limited editing capabilities. 

The problem in providing fine-grained geometric control lies in the GS representation, which is made up of an unstructured array of different 3D Gaussians whose aggregation forms visuals when splatted onto a 2D canvas. This often leads to a global dependence between different Gaussian - changes the position of one and the plausibility of the scene is ruined. Hence, it is difficult to provide the ability to perform local edits while maintaining the integrity of the resulting visuals.

To tackle these issues, in this work, we introduce the first sketch-guided 3D GS deformation system, which enables the user to intuitively interact with a simple 2D sketch of the object, and by which induces a 3D deformation of the Gaussians. To achieve this, we propose several technical contributions: 1) geometrically, to ensure the deformations produced are regulated, we propose a novel deformation framework for GS, based on cage-based deformations, which are in turn controlled by deformation Jacobians~\cite{aigerman2022neural}. 
2) Semantically, we leverage ControlNet~\cite{zhang2023adding} and Score distillation Sampling(SDS)~\cite{poole2022dreamfusion} to ensure a semantically-meaningful, plausible 3D GS deformation. Together, these two contributions enable the user to deform the shape freely, while preserving its integrity, see Figure~\ref{fig:compare}.

Our experiments verify our method's ability to provide deformation of Gaussian splats. 





\section{Related Work}
\label{sec:related_work}

\subsection{Sketch-based 3D shape Editing}
Sketching is a widely used modeling paradigm in geometric modeling and computer graphics.  
Early methods such as Teddy~\cite{igarashi2006teddy} and Fibremesh~\cite{nealen2007fibermesh} construct smooth shapes guided by user-specified 2D sketches.
After generating the initial shape, deformations can be done by drawing reference strokes or deforming the generated curve.

Some works~\cite{nealen2005sketch,kho2005sketching,zimmermann2008sketching,Kraevoy:09} focus on the sketch-based deformation that uses individual drawn strokes as a deformation clue.  (For a thorough review of the 3D shape modeling, we refer to~\cite{ding2016survey}.) 
 All these methods combine the user constraints from sketches with some geometric regularizer such as the Laplacian. Whereas, these geometric energies were designed to preserve certain properties, such as smoothness, and couldn't take into account the semantic information of the object. This can sometimes lead to unnatural deformation, e.g. bending the straight shape.   

More recently, data-driven methods were applied to sketch-based 3D shape modeling~\cite{guillard2021sketch2mesh,mikaeili2023sked,liu2024sketchdream,binninger2024sens}. Some works~\cite{guillard2021sketch2mesh,binninger2024sens} trained neural networks to generate 3D meshes conditioned on the input sketch.
These methods always need large-scale datasets to train the networks and the editing can only be done for generated shapes. 
For more general sketch-based editing, ~\cite{mikaeili2023sked,liu2024sketchdream} edited the shapes (represented by Neural Radiance Field) by 2D sketch matching and employed Score Distillation Sampling(SDS)~\cite{poole2022dreamfusion} to produce natural-looking editing that satisfied the semantic of the text prompt.  

\subsection{SDS in 3D shape Editing}
Score Distillation Sampling(SDS) was first introduced in ~\cite{poole2022dreamfusion} as a 3D shape generation method based on 2D diffusion prior. The core idea of SDS is to make the renderings of generated 3D shapes look natural from any random viewpoint. Since it is an image-based score, it can be applied to 3D editing of any representation, e.g. triangular mesh~\cite{xie2023dragd3d,yoo2024plausible}, NeRF~\cite{mikaeili2023sked} and 3D GS~\cite{ling2024align}. However, the original SDS used a 2D image diffusion model with only 2D knowledge, leading to the view inconsistency problem of SDS. 
Recently, by using 3D data in training the diffusion model, Multi-view diffusion~\cite{shi2023mvdream,liu2023zero,huang2024epidiff} was introduced to generate 3D shapes with better geometric consistency. By replacing the image diffusion model in the SDS with multi-view Diffusion, the cross-view consistency can be improved in the 3D shape editing~\cite{ren2023dreamgaussian4d,liu2024sketchdream}.

\subsection{Editing 3D Gaussian Splatting}
Neural Radiance Field(NeRF)~\cite{mildenhall2020nerf} and 3D Gaussian Splating(3D GS)~\cite{kerbl20233d} are new 3D representations designed for novel-view synthesis that the 3D scene can be reconstructed from a set of images. Some work has been done for NeRF editing~\cite{mikaeili2023sked,haque2023instruct,song2023blending,zhuang2023dreameditor}. Since our work focuses on sketch-based deformation of 3D GS, we will talk in more detail about editing 3D GS.
\cite{wang2024gaussianeditor} introduce a method that edits the 3D Gaussian Splatting(GS) scene with text instruction, powered by LLM and 2D image diffusion prior, which can achieve object texture editing and environment changing. \cite{chen2024gaussianeditor} also uses 2D image diffusion prior as the guidance of the editing but introduced the Hierarchical GS to improve the editing quality. It enables object removal and addition by employing inpainting techniques.

Instead of guiding the editing by 2D image diffusion prior,~\cite{chen2024dge,wu2024gaussctrl} introduced methods that edit the rendered image of original 3D GS from multi-views with consistency control, and fit the changes in the edited images to 3D GS directly, which improve the efficiency and quality of editing significantly.

Some works focus on the deformation of the 3D GS. Align-Your-Gaussians(AYG)~\cite{ling2024align} and DreamGaussian4D~\cite{ren2023dreamgaussian4d} introduced methods that animate a static 3D GS object to a 4D GS sequence. AYG~\cite{ling2024align} employs Video Diffusion prior to drive the temporal deformation between frames and using 2D Diffusion prior for every frame respectively to constrain the deformation validly. Instead of using Video Diffusion prior, DreamGaussian4D~\cite{ren2023dreamgaussian4d} uses a reference video in one specific view to animate the static 3D GS and applies 3D-aware Score Distillation Sampling(SDS) to propagate the deformation in every frame. The reference video was generated from an image-to-video Diffusion model based on the rendered image of static 3D GS from that specific view. Compared to AYG, DreamGaussian4D is more efficient but limited by the generation quality and universality. 

PhysGaussians~\cite{xie2024physgaussian} seamlessly integrates physically grounded Newtonian dynamics within 3D GS to achieve high-quality novel motion synthesis. It adapts the Material Point Method(MPM) to 3D GS which enriches 3D GS with meaningful kinematic deformation and mechanical stress attributes.

SuGaR~\cite{guedon2024sugar} employs new training energies based on original 3D GS~\cite{kerbl20233d} that produce 3D GS with better surface alignment and more even density distribution. These new properties make it possible to extract mesh from 3D GS only using traditional methods, such as Poisson reconstruction~\cite{kazhdan2006poisson}. Given the extracted mesh, The 3D GS can be bound to the mesh and deformed by mesh deformation algorithms, such as ARAP~\cite{sorkine2007rigid}. Similar to SuGaR,~\cite{gao2024mesh} also proposes to bind the 3D GS kernels to the mesh, which is reconstructed by the existing methods from the input multi-view images directly. However, the result significantly depends on the extracted mesh, which can fail in scenes with complex geometry and transparent components. 

\begin{figure}
    \centering
    \includegraphics[width=\linewidth]{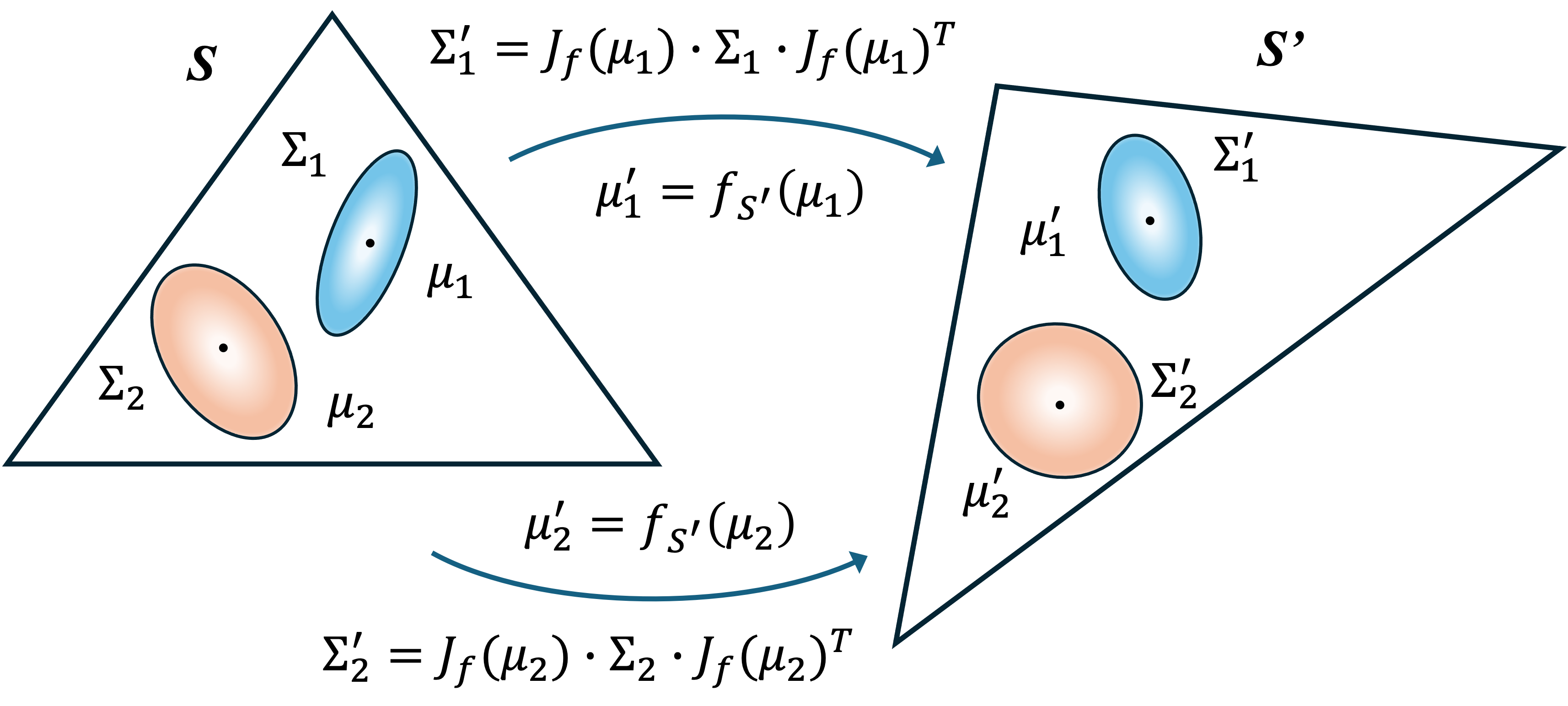}
    \caption{A simplified 2D illustration of cage-based (in this case, a triangle) deformation of Gaussian splats. $S$ and $S'$ are the original and deformed cage, respectively. $\mu_i$ and $\Sigma_i$ are the centroid and covariance of the original Gaussian splats, and $\mu_i'$ and $\Sigma_i'$ of the deformed ones. $f_S(\mu_i)$ is the cage interpolation function and $J_f(\mu_i)$ is the Jacobian matrix of the interpolation function $f$. }
    \label{fig:cbd}
\end{figure}

\section{Method}
We next detail the various components of our framework, starting with an overview of the representation of 3D Gaussians~\cite{kerbl20233d}, moving on to our cage-based deformation through jacobians, and concluding with applying this deformation technique using sketches and Score Distillation Sampling~\cite{poole2022dreamfusion}.

\begin{figure*}
     \centering
     \includegraphics[width=\linewidth]{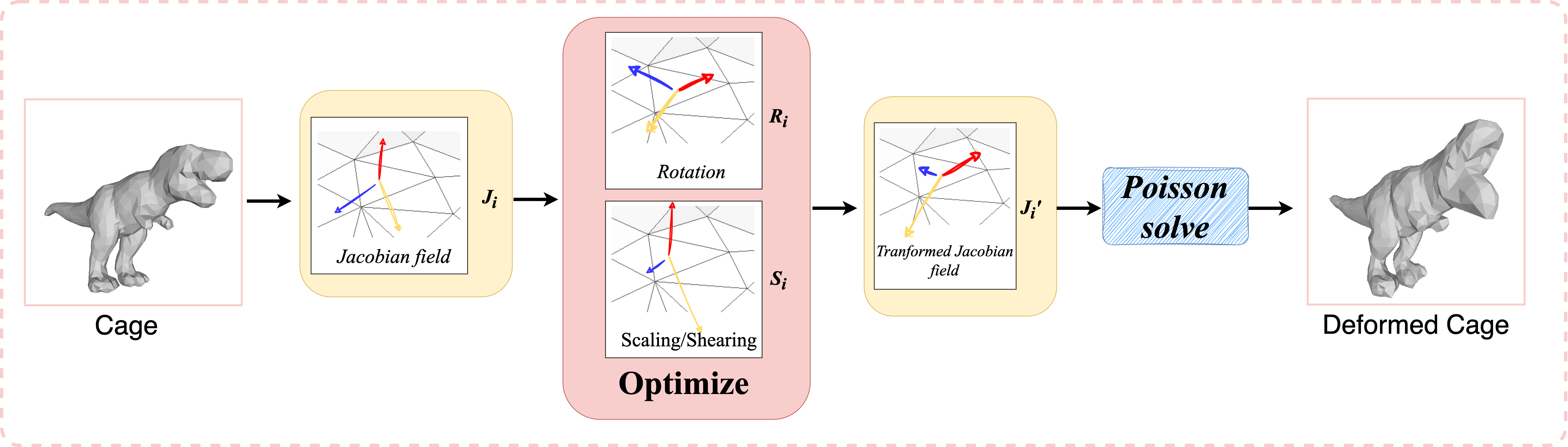}
     \caption{Controlling Cages via decomposed Neural Jacobian Fields. Instead of optimizing the NJF of the cage $J_i$, we optimized the rotation component $R_i$ and the stretch component $S_i$ of the NJF's transformation respectively. The deformed NJF can be computed easily by applying the transformation to the original NJF. Finally, the deformed cage was obtained by solving the Poisson equation. }
     \label{fig:cage_jacobian}
 \end{figure*}

\subsection{3D Gaussians}
A $3D$ Gaussian is defined by a full $3D$ matrix $\Sigma\in\mathbb{R}^{3\times 3}$ and a centroid $\mu\in\mathbb{R}^3$, defining the density of the Gaussian:
\begin{equation}
    G(x) = e^{-(1/2)(x-\mu)^T\Sigma^{-1}(x-\mu)}.
\end{equation}
Given a diagonal scaling matrix $S\in\mathbb{R}^3$ and rotation matrix $R\in SO(3)$, the corresponding $\Sigma$ is constructed as:
\begin{equation}
    \Sigma = RSS^TR^T,
\end{equation}
with $S$ and $R$ being the variables that are optimized during the reconstruction of the scene using $3D$ Gaussians. We consider a collection of such Gaussians, $\left(\Sigma_i,\mu_i\right)_{i=1}^n$.

\subsection{A Regularized Framework for Deforming 3D Gaussians}

Deforming 3D Gaussians entails assigning a new centroid position $\mu$ and a new covariance matrix $\Sigma$ to each Gaussian. However, we wish to regulate the space of possible deformations, to avoid Gaussians floating apart, and exposing only meaningful deformations of the object the Gaussians represent. Towards this goal, we design a novel, tailor-made deformation scheme, incorporating two components: 1) a cage-based deformation~\cite{ben2009variational}, tailored to Gaussian Splats; 2) a method to control this cage deformation, inspired by neural jacobian fields~\cite{aigerman2022neural}. We detail these two components next.
\label{sec:cage}
\subsubsection{Cage-Based Deformation of Gaussians}
A cage-based deformation uses a triangular mesh $S$ with vertices $V$ and triangles $T$. The mesh is deformed into another state, $S'$, by moving its vertices. By that, the mesh defines a deformation $f_{S'}:\mathbb{R}^3\to\mathbb{R}^3$, mapping every point in $\mathbb{R}^3$ as a function of the positioning of the vertices of $S'$. This enables exposing a more meaningful, low-dimensional deformation space, controlled by the cage's vertices.
There are many different approaches to define this function; we choose to use~\cite{ben2009variational} - see the supplementary material for the full details.




We next define the deformation of the Gaussians w.r.t the cage's deformation: the deformed position $\mu'$ of each centroid $\mu$ is defined as mapping it through the cage deformation: 
\begin{equation}
    \mu' = f_{S'}(\mu).
    \label{eq:cage_interpolation}
\end{equation}
Similarly, to modify the covariance matrix $\Sigma$, as is standard, we use a local linear approximation of the deformation, via the Jacobian matrix $J_f = \nabla f_{S'}$, evaluated at the centroid $\mu$. Then, the deformed covariance matrix $\Sigma'$ can be expressed as:
\begin{equation}
    \Sigma' = J_fRSS^TR^TJ_f^T.
    \label{eq:cage_interpolation_deriv}
\end{equation}

\subsubsection{Controlling Cages via Neural Jacobian Fields}

\begin{figure*}
    \centering
    \includegraphics[width=\linewidth]{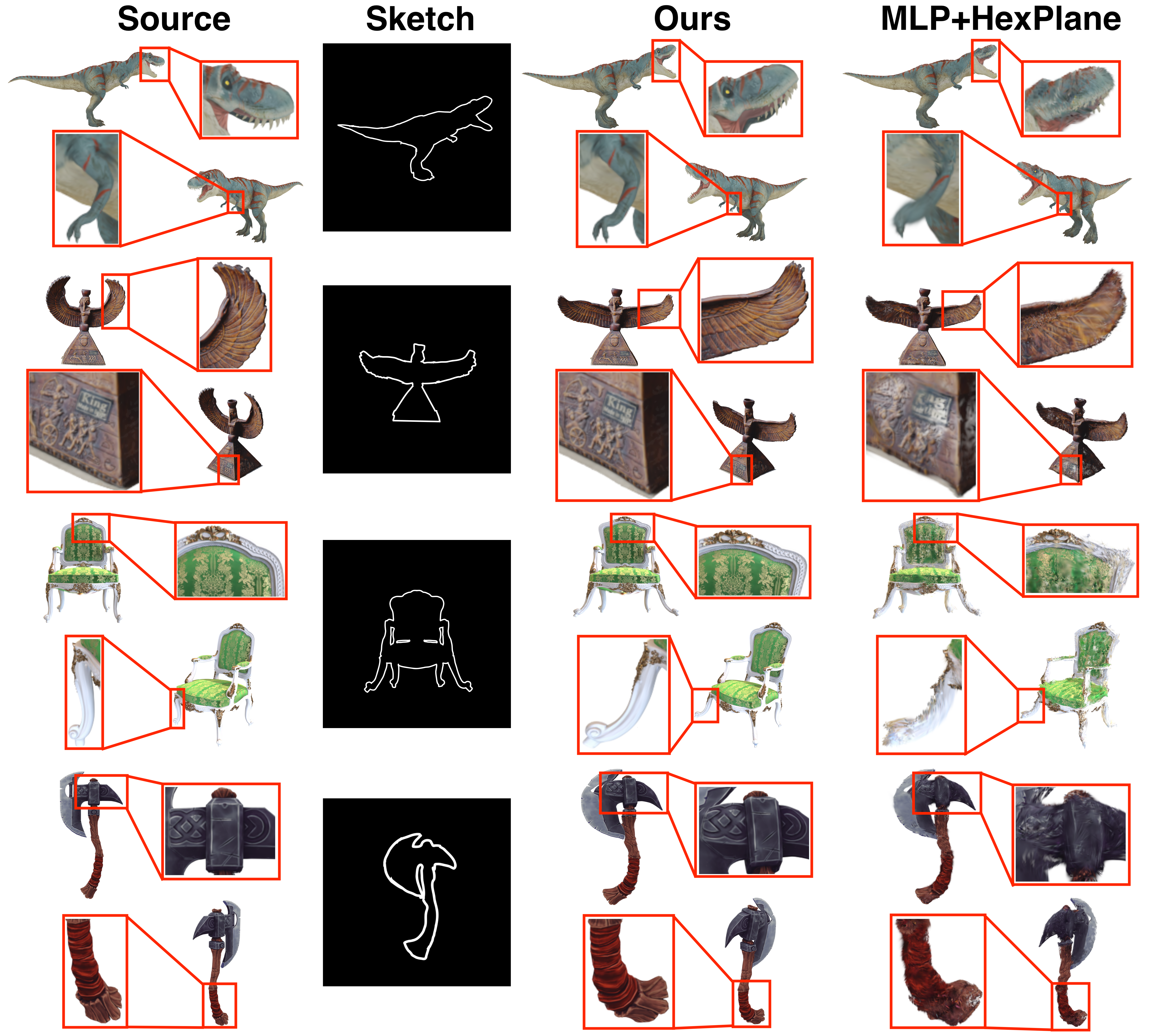}
    \caption{Comparison of using Cage or MLP+HexPlane~\cite{ren2023dreamgaussian4d} to represent deformation of the $3D$ Gaussians. MLP+Hexplane can cause severe fuzzy rendering of the deformed 3D GS because of the lack of geometry regularization. Our cage-based method produces almost lossless deformation regarding the rendering visual quality.}
    \label{fig:compare}
\end{figure*}

\label{sec:dNJF}
While the cage-based deformation already regularizes the deformation of the Gaussians, We found that optimizing directly on the cage vertices tends to produce entanglement and unsmooth cage which can lead to artifacts in the rendering of $3D$ GS, as in figure ~\ref{fig:ablation_njf}. We thus control the cage vertices' position using  
Neural Jacobian fields (NJF)~\cite{aigerman2022neural}. 
In short, NJF positions a mesh's vertices $V'$ from given per-triangle matrices  $M_i \in \Re^{3\times3}$, by minimizing the squared error between those matrices and the mesh's per-face \textit{Jacobians} $J_i \in \Re^{3\times3} $, defined as 
\begin{equation}
    J_i = V' \nabla^T_i,
\end{equation}
where $\nabla^T_i$ is the gradient operator of triangle $t_i$. The solution to this least-squares problem is achieved via \emph{Poisson's equation}, amounting to solving a single sparse linear system, which is easily implementable in a differentiable pipeline. 


We represent jacobians in a manner better accommodating for optimization: Suppose the initial per-face Jacobian is $J_i \in \Re^{3\times3}$ for face $i$, and the deformed per-face Jacobian is $J_i^{'}$. There is a linear transformation $T \in \Re^{3\times3}$ s.t. $J_i^{'} = TJ_i$. By polar decomposition, this transformation matrix can be decomposed to 
 \begin{equation}
     T = R\cdot S,
 \end{equation}
where $R$ is an orthogonal matrix and $S$ is the stretching component (symmetric semi-positive definite matrix) of the transformation.

As shown in Figure \ref{fig:cage_jacobian}, we express the deformation of the cage as a per-triangle rotation and stretching in the Jacobian space. After the cage was deformed, the deformation of $3D$ GS was computed by equations \ref{eq:cage_interpolation} and \ref{eq:cage_interpolation_deriv}.

We represent the rotational component using the smooth 6-DoF representation illustrated in~\cite{zhou2019continuity} and the stretching component as a symmetric 3 by 3 matrix with 6 degrees of freedom. Since we actually have more freedom after decomposing the Jacobian field for optimization, we found that it can produce lower energy values than optimizing the Jacobian field directly.

\subsection{3D-aware Score Distillation Sampling}

We use Score Distillation Sampling(SDS)~\cite{poole2022dreamfusion} to guide our deformation to be plausible. In short, SDS renders a 3D model from different view points, and then leverages a trained 2D image diffusion model, and backpropagates the diffusion process through the differentiable renderer, to the degrees of freedom of the 3D model. we further make use of a 3D-aware image diffusion model~\cite{liu2023zero} to enable a more accurate 3D consistency of the generated images. In short, this model can be conditioned on a specific viewing direction, and produces more consistent images of the same object from different view point. See Supplemental for the full details.


\subsection{Sketch-guided 3D GS Deformer}
The overview of the editing pipeline is shown in Figure \ref{fig:method_overview}. The user first selects a viewpoint $vp$ (blue camera). This viewpoint is used to extract a silhouette of the object. We render the $3D$ GS from viewpoint $vp$ to get an image $I_{vp}$ (Figure~\ref{fig:method_overview} B). The user additionally deforms the silhouette into $S_{vp}$ (Figure~\ref{fig:method_overview} C). To obtain the deformed image $I^{CN}_{vp}$ (Figure~\ref{fig:method_overview} D) guided by the user's sketch, the rendering $I_{vp}$ is fed into an image-to-image diffusion model conditioned on the $S_{vp}$ by using ControlNet~\cite{zhang2023adding}. Our loss measures the silhouette difference $\mathcal{L}_{sil}$ between $I^{CN}_{vp}$ and $I^{def}_{vp}$ (Figure~\ref{fig:method_overview} E), the rendering of deformed 3D GS from viewpoint $vp$:
\begin{equation}
    \mathcal{L}_{sil} = \|I^{CN}_{vp}-I^{def}_{vp}\|_2^2.
\end{equation}
We chose to only penalize the silhouette and not the full RGB deformed image, as our experiments showed the texture of the objects can be otherwise changed drastically.

To keep the deformation natural in all views, we apply 3D-aware SDS on randomly sampled views (red camera) in every iteration. Thus the final gradient used during optimization is:
\begin{equation}
\label{eq:loss_total}
    \nabla \mathcal{L}_{total} = \alpha \nabla \mathcal{L}_{sil}+\nabla \mathcal{L}_{SDS},
\end{equation}
$\alpha$ is set to $10000$ for all examples. When optimizing the 3D GS by the objective function \ref{eq:loss_total}, the 3D GS is deformed by the differentiable Cage-based block as shown in section \ref{sec:cage}.

\subsection{Implementation details}
The cage is generated automatically by first extracting a mesh from the GS model using the coarse stage of the SuGaR~\cite{guedon2024sugar} method. This mesh is very large, sometimes it is not a closed manifold and it has many fold-overs so it would not be suited for a cage. Therefore, we compute a triangulated offset surface using a function from Libigl~\cite{jacobson2013libigl} that applies marching cubes on a grid of signed distance values from the input triangle mesh.  The resulting mesh is a close manifold triangular mesh with an adjustable number of vertices depending on the model size and level of detail desired. 
We used StableDiffusion-XL as the diffusion model of the ControlNet. For each deformation, we optimized for $2000$ iterations, with a learning rate of $0.002$, and optimized by Adam optimizer~\cite{kingma2014adam}.  Except for the reference view, $4$ random views were sampled for the 3D-aware SDS. The diffusion model used in 3D-aware SDS is zero-1-to-3 XL~\cite{liu2023zero}. All experiments were run on a single Nvidia RTX A6000 GPU. The running time was related to the number of Gaussian splats and cage resolution in different scenes. For a scene with 268k gaussians and 376 cage vertices, the running time is 12 minutes. For a scene with 90k gaussians and 1343 cage vertices, the running time is 9 minutes. Thus, the running time is primarily influenced by the number of Gaussians. For large-scale scenes containing more Gaussians (e.g., 500k), the computation reaches a bottleneck, resulting in slower performance (around 40 minutes). More efficient diffusion models is a growing area of research and our method would be directly accelerated by the many methods that are being developed~\cite{starodubcev2024your,hong2024sfddm}.

\section{Experiments}

\label{sec:exp}
\subsection{Results}
\begin{figure}
    \centering
    \includegraphics[width=\linewidth]{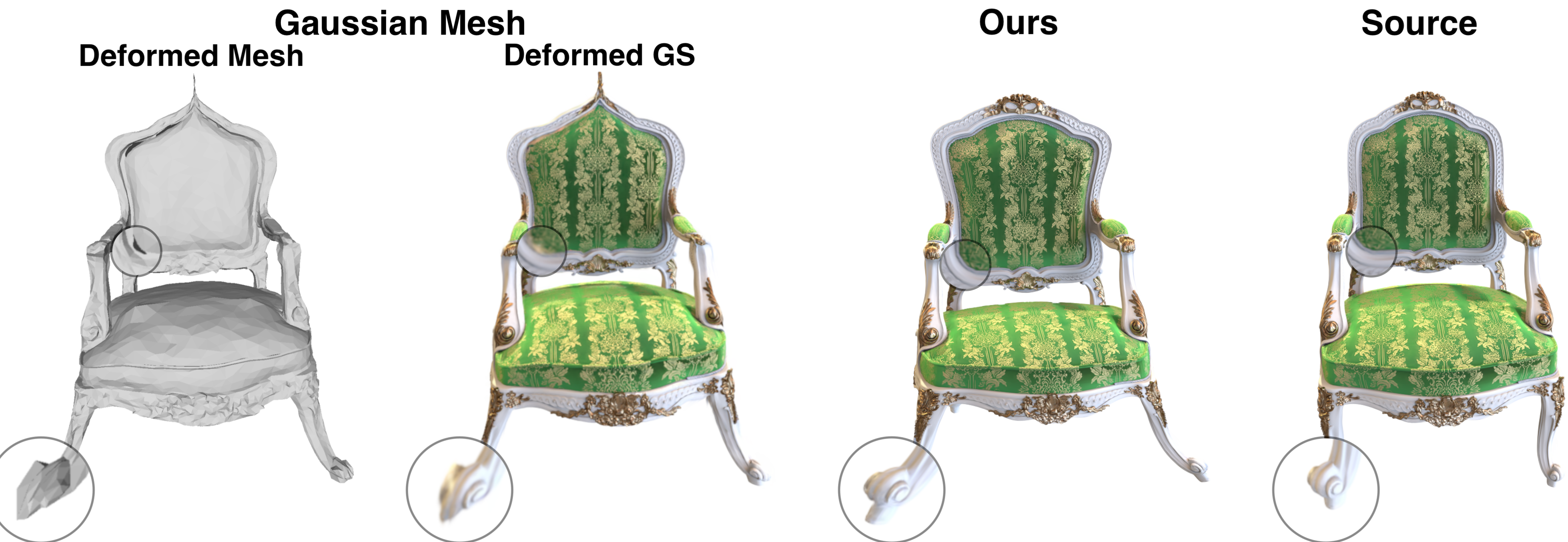}
    \caption{The comparison of our method with mesh-binding method GaussianMesh~\cite{gao2024mesh}. From left to right: 1) the deformed proxy mesh obtained by our sketch-guided pipeline for GaussianMesh. 2) the deformed 3D GS by GaussianMesh. 3) deformation result obtained by our cage-based method. 4) undeformed original 3D GS.}
    \label{fig:compare_gm}
\end{figure}

\begin{figure}
    \centering
    \includegraphics[width=\linewidth]{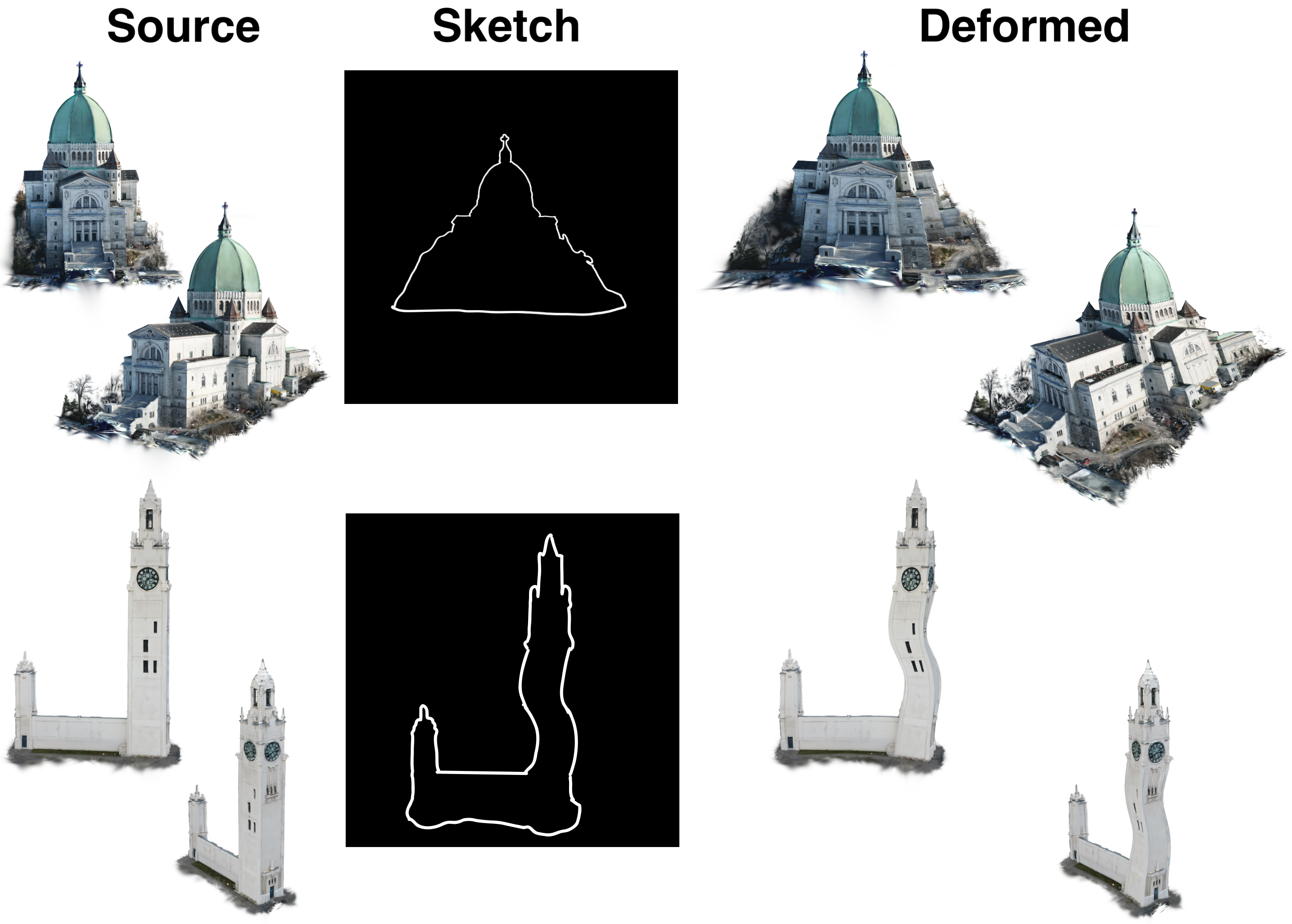}
    \caption{Deforming UAV-captured real-world large scene. We converted the UAV-captured images of a great Oratory and a clock tower into 3D GS scenes and conducted experiments, as shown in the top and bottom rows respectively.  }
    \label{fig:heritage}
\end{figure}

\begin{figure}
    \centering
    \includegraphics[width=\linewidth]{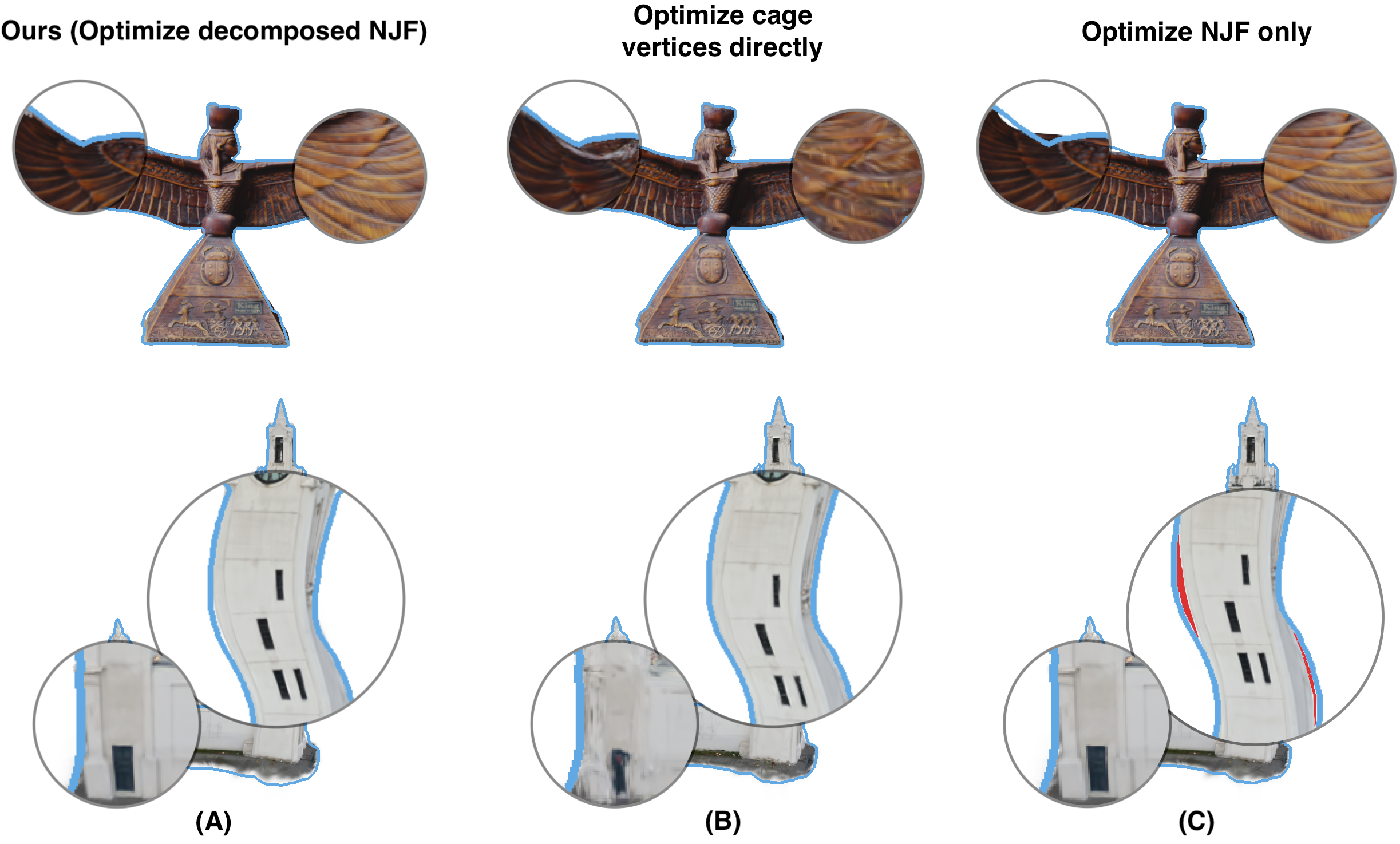}
    \caption{Ablation study on cage optimization. The blue line is the target silhouette and the the gap area between the target silhouette and the actual silhouette is marked in red. (A) Our final result uses our proposed modifications of NJF. (B) Optimizing the cage vertices directly. (C) Optimizing the original Neural Jacobian Field (without our modifications)  
    }
    \label{fig:ablation_njf}
\end{figure}
\begin{figure*}
    \centering
    \includegraphics[width=\linewidth]{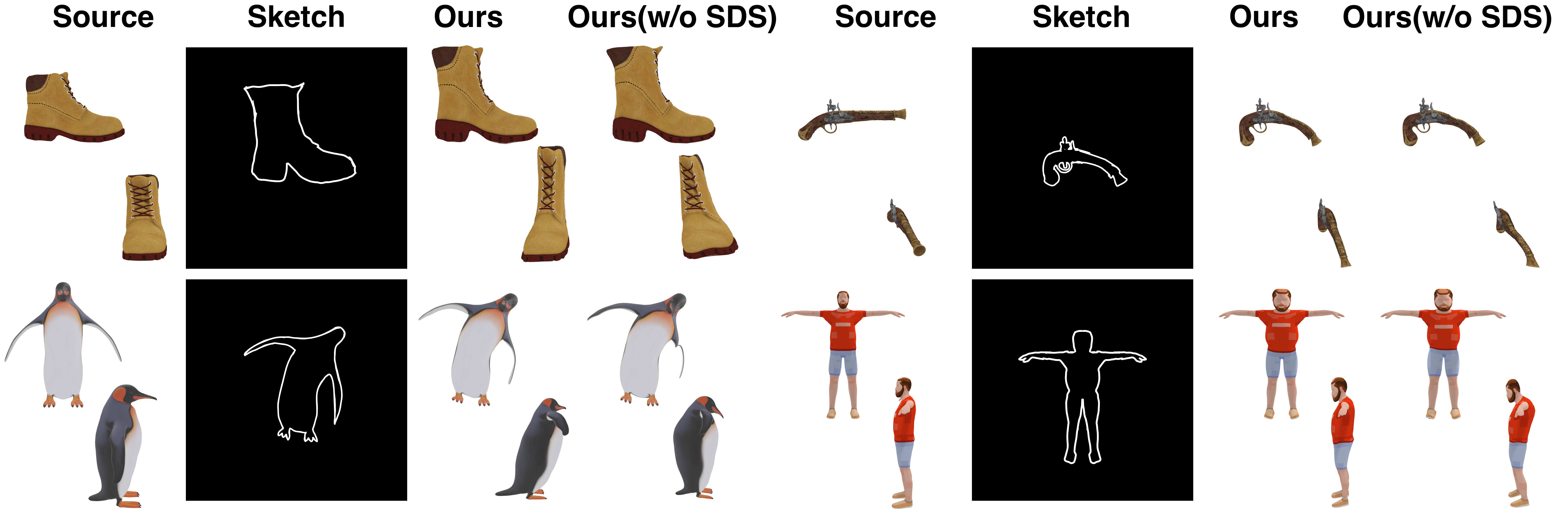}
    \caption{Ablation study of 3D-aware SDS. 3D-aware SDS was used to produce natural-looking deformation in all views. The result shows undesired distortion without 3D-aware SDS.}
    \label{fig:ablation}
\end{figure*}

We tested our method on various 3D objects, from human-made objects to animals and humans, as shown in Figure \ref{fig:teaser}, \ref{fig:compare}, and \ref{fig:ablation}. 
It shows that our method can deform the objects precisely with the guidance of the sketch and produce natural-looking results. These 3D shapes were collected from the SketchFab and TurboSquid and transferred into a 3D GS scene. The 3D shapes were originally mesh and converted into 3D GS by training from 100 sampling images from random views of the shape.
Moreover, we also explored the ability of our method to deform large-scale real-world captured data. 
We used the UAV-captured dataset of a great Oratory and a Clock Tower to reconstruct the 3D GS scene respectively. As shown in Figure \ref{fig:heritage}. 

As one of the important applications of 3D deformation is animation, we also conducted experiments of animating static 3D GS by our method.
As shown in Figure~\ref{fig:teaser}(C), and the accompanying video, the user can provide multiple input sketches as some key-frame sketches in the animation sequence. Then, by running our method for each key frame and interpolating between the deformed cage of key frames, we can get animations of the static 3D GS.

\begin{table}[]
\centering
\caption{Quantitative comparison: we report the relative CLIP-IQA score~\cite{wang2023exploring},  and relative Q-align score\cite{wu2023q}.}
\label{Quantitative_comparison}
\begin{tabular}{lll}
\hline
\multicolumn{1}{c}{} & \multicolumn{1}{c}{MLP+HexPlane} & \multicolumn{1}{c}{Ours} \\ \hline
Relative CLIP-IQA $\uparrow$ & 75.32\% & 99.03\% \\ 
Relative Q-align $\uparrow$  & 78.11\% & 99.88\%
\end{tabular}
\end{table}


We first compared our Cage-based deformation for 3D GS with the SOTA method MLP+HexPlane~\cite{ren2023dreamgaussian4d}. 
For comparison, we deform the source 3D GS using our pipeline and render the sketch view of the deformed 3D GS as the reference image of the MLP+Hexplane. 
Thus, except for the difference in the deformation method, MLP+Hexplane also has extra RGB image guidance in the sketch view when our pipeline is only guided by silhouette. Even in that case, our pipeline can deform the 3D GS with much higher fidelity than using MLP+Hexplane. The qualitative comparison was shown in Figure \ref{fig:compare}, because of the good space continuity of the cage deformation, the local detail features can be fully preserved, e.g. the pattern on the statue, and the teeth of the dinosaur. Though the Hexplane was applied as a geometric regularizer, the detailed features can still be destroyed and cause fuzzy renderings when using MLP+Hexplane. We quantitatively compare the visual quality of the deformed shapes based on image quality. We use the CLIP-IQA~\cite{wang2023exploring}, a metric measuring the image quality based on the CLIP score, and Q-align\cite{wu2023q}, an image quality assessment based on large multi-modality models (mPLUG-Owl2~\cite{ye2024mplug}), to evaluate 8 deformed results. For every deformed 3D GS, 8 views were rendered as the images to be assessed. Since our task is to measure how the image quality was changed after the deformation process, we report the relative score of both CLIP-IQA and Q-align, which is calculated as the metric score of the undeformed renderings divided by the metric score of the deformed renderings. As shown in table~\ref{Quantitative_comparison}, the average decreases of the CLIP-IQA score and Q-align score for the MLP+Hexplane method are $24.68\%$ and $21.89$ respectively, however, our method almost keeps the same CLIP-IQA and Q-align score as original renderings with less than $1\%$ decreasing. We also conducted a user study with 15 participants comparing the deformation fidelity to using MLP+HexPlane across 6 examples with instruction: which deformed result has better visual fidelity?. Among these 90 comparisons, our method gains $90.0\% $ preference from the participants. 

We also include a comparison with mesh-binding GS (GaussianMesh~\cite{gao2024mesh}); however, a direct substitution of our cage-based method with their method is not straightforward, as their deformation approach is not implemented as a differentiable process. We perform the comparison with the following setup: 1) Run the mesh reconstruction method~\cite{instant-nsr-pl} to obtain a proxy mesh $M_p$, which is the base mesh to reconstruct the 3D GS. 2) Run GaussianMesh~\cite{gao2024mesh} reconstruction with input $M_p$ to get the 3D GS. 3) Run our pipeline as shown in figure~\ref{fig:method_overview} in which the 3D GS was replaced by $M_p$  and regularized by NJF to get a deformed mesh $M'_p$. 4) Run GaussianMesh Deformation based on the deformed mesh $M'_p$ to get the deformed 3D GS. The result is shown in figure~\ref{fig:compare_gm}. The deformed mesh exhibits severe artifacts, including entanglement and faces inversion, even with the NJF regularization. These artifacts are reflected in the final rendering of the 3D GS. Another drawback of the mesh-binding method is that the reconstruction is based on a reconstructed mesh obtained by other methods~\cite{instant-nsr-pl}, which can fail in complex scenes. We show an example of a reconstructed mesh in the Supplementary material.

Additionally, we evaluated the silhouette adherence by the Intersection of Union (IoU) between the alpha rendering of 3D GS and the target silhouette from the sketch view. Among the examples, the average IoU before deformation is $0.654$ and after deformation is $0.901$.

\subsection{Ablation}
We tested the effects of two important components in our method, decomposed NJF and 3D-aware SDS. 
We explored the difference of optimizing directly on the cage vertices, via NJF and our proposed decomposed NJF. As shown in Figure~\ref{fig:ablation_njf}, optimizing directly on the cage vertices can lead to undesired fuzzy rendering, which is caused by entanglement of the cage during the optimization. As shown in Figure~\ref{fig:ablation_njf} (B), the detailed feather of the statue and the door on the side of the clock tower were destroyed. Applying NJF can solve this problem since it works as a geometry regularizer for the cage. However, it can be too strong to fit the sketched silhouette precisely. As shown in Figure~\ref{fig:ablation_njf} (C), the wing of the statue is not fitted on the end and the clock tower is not bent enough. When using decomposed NJF as stated in section \ref{sec:dNJF}, we can obtain a more efficient optimization which leads to a lower $\mathcal{L}_{sil}$ (average of  $49.36\%$ decreasing in these two examples), but with same rendering visual quality compared to NJF (Figure~\ref{fig:ablation_njf} A).

We also explored the effect of 3D-aware SDS in our pipeline. As shown in Figure \ref{fig:ablation}, this plays a critical role in preserving undesirable deformation in the views except for the sketch view in the pipeline. For instance, when the pistol was deformed only by $\mathcal{L}_{sil}$, the barrel was bent, which can be fixed by adding the 3D-aware SDS.

\subsection{Failure cases and mitigations}
Some failures can occur from single view mismatches (e.g. mismatching of left and right legs in a side view). This can be addressed by slightly changing the view (See supplementary material) 

\section{Conclusion}
In this work, we propose a novel sketch-guided 3D Gaussian Splatting deformation framework, which enables intuitive, fine-grained control over the geometry of 3D GS by a single-view silhouette sketch. 
Geometrically, we designed a novel cage-based deformation tailored to Gaussian Splats and optimized its position using a modified Neural Jacobian Fields formulation.
As the rendering of Gaussian Splats overlays intersecting splats on top of each other, deforming these splats can lead to visual artifacts due to misalignment. Our method provides accurate alignment of the splats in the deformed pose that yields crisp rendered results that adhere closely to the input sketch. 
To ensure semantically meaningful deformation from any viewpoint, our method leverages ControlNet as well as Score distillation sampling.
\section{Limitation and Future work}
    First, our method relies on ControlNet and an image-to-image diffusion model to translate the user-drawn silhouette sketch into a usable constraint for the deformation system. However, the ability of ControlNet to generate a reference image of the deformed object is sometimes limited.    
    Second, although the cage for 3D GS is generated automatically, the process of training the SuGaR model to extract a mesh from the 3D GS still takes several minutes, which limits the system's efficiency.

In future work, we will focus on improving the reliability of the sketch-guided deformation, particularly by enhancing the quality and consistency of the ControlNet-generated reference images. Additionally, making the cage generation process more efficient could enhance the practicality of our method. Exploring these avenues would contribute to both the robustness and efficiency of the system.

\section{Acknowledgment}
We acknowledge the support of the Natural Sciences and Engineering Research Council of Canada (NSERC), under funding reference numbers RGPIN-2021-03477 and RGPIN-2024-04605, as well as the support of Fonds de recherche du Québec – Nature et technologies (FRQNT), under funding reference number 365040.
\vspace{-3em}
{
    \small
    \bibliographystyle{ieeenat_fullname}
    \bibliography{main}
}

\end{document}